\documentclass[11pt,a4paper]{article}
\usepackage[hyperref]{emnlp2018}
\usepackage{times}
\usepackage{latexsym}
\usepackage{times}
\usepackage{helvet}
\usepackage{courier}
\usepackage{color,soul}
\usepackage[english]{babel}
\usepackage{graphicx}
\usepackage{subcaption}
\usepackage{hhline}
\usepackage{pbox}
\usepackage{float}
\usepackage{amsfonts, amsmath}
\usepackage{url}
\usepackage{array}

\aclfinalcopy 

\title{Dropout during inference as a model for neurological degeneration in an image captioning network}

\author{Bai Li$^{1, 2}$ \quad Ran Zhang$^1$ \quad Frank Rudzicz$^{3, 1, 2}$ \\
  $^1$University of Toronto, $^2$Vector Institute, $^3$Toronto Rehabilitation Institute\\
  {\tt \{bai,ranzhang,frank\}@cs.toronto.edu}}

\date{}

\begin{document}
\maketitle
\begin{abstract}
We replicate a variation of the image captioning architecture by \citet{show-and-tell}, then introduce dropout during inference mode to simulate the effects of neurodegenerative diseases like Alzheimer's disease (AD) and Wernicke's aphasia (WA). We evaluate the effects of dropout on language production by measuring the KL-divergence of word frequency distributions and other linguistic metrics as dropout is added. We find that the generated sentences most closely approximate the word frequency distribution of the training corpus when using a moderate dropout of 0.4 during inference.
\end{abstract}

\section{Introduction}

Visual understanding and language generation are two tasks that are intuitive for humans, but pose a challenge to computers. Recently, convolutional neural networks (CNNs) \citep{cnn-imagenet} and long short-term memory networks (LSTMs) \citep{lstm} have achieved state-of-the-art results on image understanding and natural language generation, respectively. For image captioning, one of the most successful approaches has been the encoder-decoder architecture, where the image is first ``encoded'' by a CNN into a latent semantic hidden vector, and then ``decoded'' into a natural language sentence using a recurrent, language generating LSTM.

For most people, describing a picture is an intuitive task that requires little effort. However, patients with neurodegenerative disorders have impaired brain function that inhibits some cognitive subtasks; this manifests as difficulties with picture description \citep{picture-description-alzheimer}. In fact, because linguistic impairment is one of the earliest signs of AD, picture description is useful as a cognitive test for AD \citep{detecting-alzheimer-picture}.

In this paper, we implement a variant of the ``Show and Tell'' neural network for image captioning \citep{show-and-tell}, and simulate the effects of neurodegeneration by adding dropout during inference, which randomly sets a subset of the neuron outputs in a layer to zero. We evaluate the effects of dropout on language generation, and compare the results to picture description by patients with diseases like AD and aphasia.

\section{Related Work}

\subsection{Image Captioning}

\citet{mao2014} were the first to apply deep recurrent neural networks (RNNs) to image captioning. Their architecture used a multimodal layer that combined image representations preprocessed by a CNN, the input word embedding, and the output of the RNN at each time step to generate output word embeddings.

\citet{show-and-tell} used an encoder-decoder architecture that only looked at the image once. The encoder was a CNN that encoded images into vector representations and the decoder was an LSTM that decoded the image features into natural language descriptions. By using an LSTM, their model was able to retain long-term dependencies and avoid having to show the image to the RNN multiple times. \citet{show-attend-tell} added a visual attention mechanism, which learned which part of the image to focus on at each time step.

\subsection{Picture Description}

Alzheimer's disease (AD) is a neurodegenerative disorder affecting 47 million people worldwide \citep{alzheimer-is-bad}. One of the earliest symptoms of AD is cognitive impairment, especially difficulty with language production. One widely-used cognitive test for AD is the {\em Cookie Theft} picture description task from the Boston Diagnostic Aphasia Examination \citep{boston-exam}. In this task, the patient is shown a drawing of a chaotic kitchen scene, and is asked to describe it in as much detail as possible. Language in patients with AD is characterized by semantic impairment, particularly difficulty finding words for concepts and ideas \citep{alzheimer-language-review}.

Wernicke's aphasia (WA), also known as fluent aphasia, is caused by damage to Wernicke's area, which is partially responsible for language production. Patients with WA tend to produce long stretches of words in a seemingly random order\footnote{Video of a patient with WA performing a picture description task: \url{https://www.youtube.com/watch?v=xzp-XUBknQI}} \citep{buckingham-fluent-aphasia}.

Recently, methods have been developed for computational methods to diagnose AD automatically from speech. For example, \citet{fraser2016} extracted a wide range of linguistic features to classify AD with about 82 percent accuracy, given a few minutes of speech from picture description.

\subsection{Dropout in neural networks}

Dropout is the technique of randomly selecting, with probability $p$, neurons in a layer and setting their outputs to zero. When used during training, dropout has been shown to be an effective regularization method and has an effect similar to averaging an ensemble of models \citep{dropout}. Dropout has been applied to RNN language models as well, with a similar regularizing effect \citep{zaremba-rnn-dropout}. While previous work considered dropout during training as a regularization mechanism, we consider dropout during {\em inference} and its effects on language generation. To our knowledge, dropout during inference in RNNs has not been studied.

AD may be caused by a misfolding in the beta-amyloid protein, causing beta-amyloid plaques to form in neurons \citep{century-of-alzheimer}. This causes inhibited electrochemical signal transmission in the synapses. Thus, the effect of AD in brain cells is similar to that of dropout in neural networks. The goal of our work is to simulate the effects of AD to produce pathological linguistic effects similar to picture description by patients with neurodegenerative disorders.

\section{Model}

\subsection{Image caption network}

We implement an encoder-decoder neural architecture. For the CNN, we use the VGG16 convolutional network \citep{vggnet} and initialize it with weights from Torchvision \cite{pytorch}, which were trained on ImageNet. During the training of our image captioning network, all of the weights for the CNN are frozen.

During inference, the image is processed by the VGG16 network up until the last hidden layer. This hidden layer is fed into a linear layer to produce a representation for the initial hidden layer of the LSTM. Then, a sentence is generated as follows: On each iteration, the LSTM hidden layer is used to produce a linear softmax classification to generate a probability distribution over the entire vocabulary to pick the next word. The word with the highest probability is picked as the next word in the sentence. We then use a word embedding lookup to convert this word into a vector, and feed it into the next iteration of the LSTM. This process continues until the LSTM generates a special end marker, or exceeds a fixed word limit of 20 words (fewer than 3\% of sentences in the training data exceed this length).

To train the network, we compute the {\em perplexity}, which represents the likelihood function of this image-caption pair according to our model, and is computed by feeding the image and caption sentence into the network and taking the sum of the cross entropy errors at each step. Once computed, the perplexity is minimized using backpropagation and stochastic gradient descent.

\subsection{Dropout in GRU}

The LSTM was the first recurrent neural network model that addressed the vanishing and exploding gradient problems, in which simple RNNs had difficulty learning long-term dependencies due to numerical instability \cite{bengio-rnn-difficult}. More recently, the gated recurrent unit (GRU) was found to achieve similar performance to the LSTM, while using only a hidden state and no cell state \citep{cho-gru}. We use a modified version of the GRU, in which a constant level of dropout is added to the hidden state between iterations. This is represented by the following equations:

$$r = \sigma(x_t U_r + h_{t-1} W_r + b_r)$$
$$z = \sigma(x_t U_z + h_{t-1} W_z + b_z)$$
$$\bar{h} = \tanh(x_t U_{\bar{h}} + (r \times h_{t-1}) W_{\bar{h}} + b_{\bar{h}})$$
$$h_t = \mathrm{dropout}(z \times h_{t-1} + (1-z) \times \bar{h})$$
where $\times$ denotes element-wise multiplication between vectors and $\sigma$ is the sigmoid. The variable $h_{t}$ is the hidden state at time $t$, and $x_t$ is the input vector. First, we compute vectors $r$ representing the {\em reset} gate, and $z$ representing the {\em update} gate, both in the range $[0,1]$. A temporary hidden state $\bar{h}$ is computed from the input $x_t$ and previous hidden state $h_{t-1}$ masked by the reset gate $r$. The next hidden state $h_t$ is set to a combination of the previous hidden state $h_{t-1}$ and the temporary hidden state $\bar{h}$ as determined by the update gate $z$. Finally, a dropout is applied to $h_t$. The variables $U$, $W$ and $b$ (with various subscripts) are unknown weights to be learned by backpropagation.

We define $d_t$ to be the dropout probability during training, and $d_e$ the dropout probability during evaluation. Typically, dropout is used during training for regularization and turned off during evaluation, so $d_t > 0$ and $d_e = 0$. However, in this work we explore the possibility of $d_e > 0$ and its effects on language generation.

\subsection{Evaluation metrics}

To test the performance of our model, we generate captions for images in a validation set, and consider the BLEU and METEOR scores; for BLEU, we consider $n=4$, which correlates most highly with human ratings of performance \citep{bleu-metric,meteor-metric}.

Next, we evaluate the effects of dropout on caption accuracy and vocabulary diversity. We use Kullback-Leibler (KL) divergence to measure the distance between the word frequency distribution of the ground-truth validation captions and the sentences generated by our network on the validation set. That is, $$D_{KL}(P || Q) = \sum_w P(w) \log \frac{P(w)}{Q(w)},$$ where $P$ is the word distribution of the generated captions and $Q$ is the word distribution of the most common 10,000 words among the ground-truth captions.

For each run of the experiment, we also calculate $|V|$, the number of unique words among all generated captions, and $p(len > 20)$, the proportion of generated captions that exceed the word limit of 20.

\section{Results and Discussion}

We train our model using the COCO2014 dataset \citep{mscoco}, which contains 82,783 training images and 40,504 validation images. 
\begin{table*}
\centering
\caption{Effects of $d_t$ and $d_e$ on caption accuracy and vocabulary diversity.}
\begin{tabular}{|c|c|c|c|c|c|c|}
\hline
$d_t$ & $d_e$ & BLEU-4 & METEOR & $D_{KL}$ & $|V|$ & $p(len > 20)$ \\
\hline
0.0 & 0.0 & 20.1 & {\bf 19.8} & 0.453 & 733 & 0.00 \\
\hline
0.0 & 0.2 & 16.5 & 18.4 & 0.298 & 2158 & 0.02 \\
\hline
0.0 & 0.4 & 7.4 & 14.0 & 0.270 & 7500 & 0.35 \\
\hline
0.0 & 0.6 & 2.2 & 9.3 & 0.823 & 9303 & 0.78 \\
\hline
0.0 & 0.8 & 0.2 & 5.4 & 1.496 & 9585 & 0.95 \\
\hhline{|=|=|=|=|=|=|=|}
0.2 & 0.0 & {\bf 20.3} & 19.5 & 0.497 & 630 & 0.00 \\
\hline
0.2 & 0.2 & 19.0 & 19.0 & 0.409 & 1312 & 0.00 \\
\hline
0.2 & 0.4 & 15.4 & 17.3 & {\bf 0.260} & 7007 & 0.03 \\
\hline
0.2 & 0.6 & 3.3 & 9.6 & 1.837 & 9841 & 0.83 \\
\hline
0.2 & 0.8 & 0.1 & 3.0 & 3.106 & 9840 & 0.99 \\
\hline
\end{tabular}
\label{table:dropout-scores}
\end{table*}

The neural network model is implemented using PyTorch \citep{pytorch}. We use GLoVe embeddings from SpaCy \citep{glove, spacy} trained on Common Crawl. The network was trained using Adam \citep{adam}.

The LSTM version of our model achieves a BLEU-4 score of 20.6 and METEOR score of 20.0. The GRU version performed similarly, with a BLEU-4 score of 20.1 and METEOR score of 19.8.

Our model slightly underperforms the scores reported by \citet{show-and-tell}. We did not implement beam search to minimize perplexity across a sequence but instead used the greedy approach of picking the highest probability word at each step. Additional hyperparameter optimization may have improved our model accuracy, though that is not our purpose here.

Next, we evaluate the accuracy and vocabulary diversity of the model as dropout is added in inference. We train two versions of the GRU model, once with no training dropout ($d_t = 0$) and once with $d_t = 0.2$. For each model, we generate captions for the validation set, using evaluation dropouts $d_e = [0.0, 0.2, 0.4, 0.6, 0.8]$. The results are shown in Table~\ref{table:dropout-scores}.

For both versions of the model, the BLEU-4 and METEOR scores are maximized when $d_e = 0$; this is expected, since dropout is usually disabled during evaluation for best performance. When evaluation dropout is moderate ($d_e = 0.4$), the model trained with dropout performs better than the model trained without. When evaluation dropout is high ($d_e \geq 0.6$), both models perform poorly.

The standard model without dropout only generates a vocabulary of 733, out of a total possible vocabulary of 10,000; when dropout is added during inference, the generated vocabulary is more diverse. In both versions of the model, the KL divergence of word frequency distributions is minimized using a moderate dropout ($d_e = 0.4$). Thus, a moderate amount of dropout produces a word frequency distribution closer to the target distribution when dropout is added during inference. However, when evaluation dropout is too high, the generated captions have low BLEU-4 and METEOR scores, high $D_{KL}$, and high increased probability of exceeding the sentence length limit.

Next, we comment on the qualitative effects of dropout on language generation. A sample of captions generated with various levels of dropout is given in the appendix. Generally, errors follow into two common patterns:

\begin{enumerate}
\setlength{\itemsep}{0pt}
\setlength{\parsep}{0pt}
\item A caption starts out normally, then repeats the same word several times: {\em ``a small white kitten with red collar and yellow chihuahua chihuahua chihuahua''}
\item A caption starts out normally, then becomes nonsense: {\em ``a man in a baseball bat and wearing a uniform helmet and glove preparing their handles won while too frown''}
\end{enumerate}

Both of these patterns are sometimes observed in patients with paraphasic disorders, like fluent aphasia \citep{neurobiology-brain}. In particular, one symptom of Wernicke's Aphasia is the use of {\em jargon speech}: long streams of words with seemingly no meaning but retaining some phonological and grammatical structure. \citet{buckingham-fluent-aphasia} give some examples of jargon speech:

\begin{itemize}
\setlength{\itemsep}{0pt}
\setlength{\parsep}{0pt}
\item {\em ``I know a deprecol over american person churches such as no dish or penthenis''}
\item {\em ``I think my foremust acoushner looks ellington''}
\item {\em ``I would say that the mick daysas nosis or chpickters''}
\end{itemize}

\section{Conclusion and Future Work}

In this paper, we have implemented an encoder-decoder image captioning neural network, and applied dropout during inference to simulate the effects of neurodegeneration on language production. The resulting sentences qualitatively resemble speech produced by patients with language production disorders like fluent aphasia. Future research will include quantitative comparisons of our language model with speech by aphasic patients.

Our model applies dropout in the hidden layer of the GRU, but there are other ways to simulate neurodegeneration as well. For example, one may instead add Gaussian noise to the hidden layer, or deterministically dropout specific neurons instead of at random. Furthermore, it is unknown whether the effects we observed can be reproduced using different corpora. The next step is to train a recurrent neural language model on a picture description corpus, such as the DementiaBank Corpus in the TalkBank project \citep{dementiabank}.

This work may be suitable as a data augmentation preprocessing step for systems that automatically detect dementia and aphasia using speech. A neural language model is trained using normative data, then dropout is applied during inference to generate degenerate data. This synthetic data is then combined with real data from patients with neurodegenerative disorders using semi-supervised methods to train a classifier. However, the implementation of this concept is a topic for future research.

\bibliography{main}
\bibliographystyle{acl_natbib_nourl}

\appendix

\section{Example sentences}

Below are a set of examples of sentences generated with dropout during inference. The neural network is trained with $d_t = 0.2$ and evaluated with $d_e = [0.0, 0.2, 0.4, 0.6, 0.8]$. Ellipsis means the sentence exceeded the length limit of 20.

\begin{itemize}
  \item $d_t = 0.2$, $d_e = 0.0$
  \begin{itemize}
    \item {\em a bear walking through a field of tall grass}
    \item {\em a group of people standing on top of a snow covered slope}
    \item {\em a cat laying on a bed with a cat}
  \end{itemize}
  \item $d_t = 0.2$, $d_e = 0.2$
  \begin{itemize}
    \item {\em a baseball player swings his bat at a baseball}
    \item {\em a group of people sitting around a table with food}
    \item {\em a boat is in the water near a dock}
  \end{itemize}
  \item $d_t = 0.2$, $d_e = 0.4$
  \begin{itemize}
    \item {\em a man in a red white and black hair is lying on a fitting fitting}
    \item {\em a herd of sheep grazing in a field of grass}
    \item {\em a table with a lot of food including grapes melon seaweed}
  \end{itemize}
  \item $d_t = 0.2$, $d_e = 0.6$
  \begin{itemize}
    \item {\em a bathroom with a toilet ripped out}
    \item {\em a bus is parked by packed packed macbook sequential sequential drawer basebal funky western sanctioned confident automobile leaguer crossroad peson ...}
    \item {\em a herd of sheep are in a williams williams twp poeple khakis accommodate surveying unenthused unenthused homey recreation slider clinton ...}
  \end{itemize}
  \item $d_t = 0.2$, $d_e = 0.8$
  \begin{itemize}
    \item {\em professional clothing great overstuffed handlebars tailed prepped photos recovery version volkswagen brings lose broiled papered sprouting valve mets halfway lavishly ...}
    \item {\em seven shoppers hunched petite westmark gril chives caucasian end yellowed trashcans crumb photographic slipper poeple poeple soaked barbecuing twp twp ...}
    \item {\em a offspring fitted stemmed pregnant nurse urns surveys consume reservoir snuggled meatballs curry twp terrace trailers peple motocross youngsters specialized ...}
  \end{itemize}
\end{itemize}

\end{document}